\documentclass[sigconf]{acmart}
\usepackage{algorithm}
\usepackage{algpseudocode}
\usepackage[capitalise, noabbrev]{cleveref}
\usepackage{multirow}
\AtBeginDocument{%
  }

\setcopyright{acmlicensed}
\copyrightyear{2018}
\acmYear{2018}
\acmDOI{XXXXXXX.XXXXXXX}
\acmConference[Conference acronym 'XX]{Make sure to enter the correct
  conference title from your rights confirmation email}{June 03--05,
  2018}{Woodstock, NY}
\acmISBN{978-1-4503-XXXX-X/2018/06}

\acmSubmissionID{8631}

\begin{document}

\title{SImpHAR: Advancing impedance-based human activity recognition using 3D simulation and text-to-motion models}

\author{Lala Shakti Swarup Ray}
\affiliation{%
  \institution{DFKI, RPTU}
  \city{Kaiserslautern}
  \country{Germany}
  }
\email{lala_shakti_swarup.ray@dfki.de}
\orcid{0000-0002-7133-0205}

\author{Mengxi Liu}
\affiliation{%
  \institution{DFKI, RPTU}
  \city{Kaiserslautern}
  \country{Germany}}
\email{mengxi.liu@dfki.de}

\author{Deepika Gurung}
\affiliation{%
  \institution{DFKI, RPTU}
   \city{Kaiserslautern}
  \country{Germany}}
\email{deepika.gurung@dfki.de}

\author{Bo Zhou}
\affiliation{%
  \institution{DFKI, RPTU}
   \city{Kaiserslautern}
  \country{Germany}}
\email{bo.zhou@dfki.de}

\author{Sungho Suh}
\authornote{Corresponding Author}
\affiliation{%
  \institution{Korea University}
  \city{Seoul}
  \country{Republic of Korea}}
\email{Sungho_Suh@Korea.ac.kr}

\author{Paul Lukowicz}
\affiliation{%
  \institution{DFKI, RPTU}
  \city{Kaiserslautern}
  \country{Germany}}
\email{paul.lukowicz@dfki.de}

\renewcommand{\shortauthors}{Ray et al.}

\begin{abstract}
Human Activity Recognition (HAR) with wearable sensors is essential for applications in healthcare, fitness, and human-computer interaction. Bio-impedance sensing offers unique advantages for fine-grained motion capture but remains underutilized due to the scarcity of labeled data. We introduce SImpHAR, a novel framework addressing this limitation through two core contributions. First, we propose a simulation pipeline that generates realistic bio-impedance signals from 3D human meshes using shortest-path estimation, soft-body physics, and text-to-motion generation serving as a digital twin for data augmentation. Second, we design a two-stage training strategy with decoupled approach that enables broader activity coverage without requiring label-aligned synthetic data. We evaluate SImpHAR on our collected ImpAct dataset and two public benchmarks, showing consistent improvements over state-of-the-art methods, with gains of up to 22.3\% and 21.8\%, in terms of accuracy and macro F1 score, respectively. Our results highlight the promise of simulation-driven augmentation and modular training for impedance-based HAR.
\end{abstract}




\maketitle

\section{Introduction}
Human Activity Recognition (HAR) using wearable sensors is foundational in healthcare, fitness, industrial safety, and human-computer interaction~\citep{bian2019passive,zhou2016never,jannat2023efficient,suh2023worker,6910440,ray2024har}. The performance of HAR systems heavily depends on the sensing modality. While prior work has explored inertial, acoustic, field-based, and physiological sensors~\citep{bian2022state,dang2020sensor}, bio-impedance remains significantly underexplored—despite its ability to capture fine-grained physiological changes that many modalities overlook.

Bio-impedance measures the opposition of biological tissue to a small electrical current passed between electrodes on the body. This impedance changes with posture, motion, muscle activation, and tissue deformation—making it potentially powerful for capturing subtle motion dynamics. Its applications extend from body composition analysis~\citep{matthie2008bioimpedance}, respiration~\citep{lee2015integrated}, and heart rate~\citep{gonzalez2008heart,sel2021non}, to dynamic contexts like gesture recognition~\citep{liu2024iface,chen2021bio}, dietary monitoring~\citep{liu2023ieat}, and activity tracking~\citep{liu2024imove,ring2015temperature}.
Despite these promising properties, bio-impedance-based HAR suffers from two major limitations: (1) the scarcity of large-scale, labeled datasets due to hardware complexity and the need for careful electrode placement; and (2) intra-user variability caused by physiological factors such as hydration level, muscle mass, and skin contact quality. These challenges hinder reproducibility and limit generalization across users.
Meanwhile, mature sensing modalities like IMUs benefit from public benchmarks~\citep{stromback2020mm,ciliberto2021opportunity++,yoshimura2024openpack} and established data augmentation pipelines~\citep{kwon2020imutube,fortes2021translating,leng2024imugpt,zolfaghari2024sensor}. To bridge the gap, simulation-based frameworks have emerged in HAR research—using digital twins to generate synthetic sensor signals from video, pose, or text~\citep{young2011imusim,rey2024enhancing,ray2023pressuretransfernet,ray2023pressim}. However, none of these approaches address bio-impedance, primarily because: (1) it has been historically treated as noise in motion contexts, and (2) no well-established model exists to map motion dynamics to time-varying impedance.

\begin{figure}[!t]
\centering
\includegraphics[width=1\linewidth]{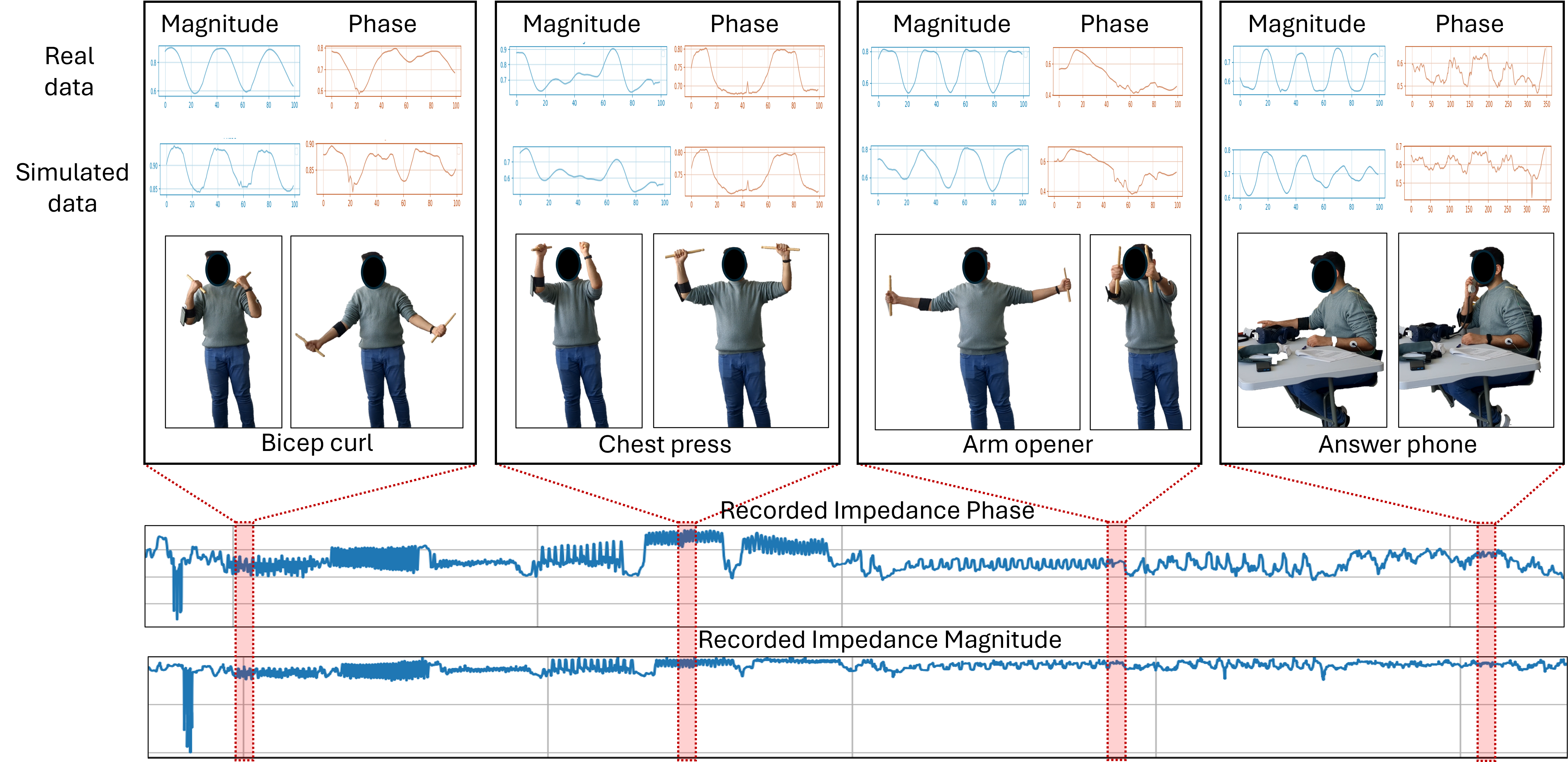}
\caption{Synthesized impedance signal (SImp) vs. real impedance signal (IMP) magnitude generated using SImpHAR.}
\label{fig:dataset}
\end{figure}

To address these gaps, we introduce \textbf{SImpHAR}, the first framework for simulating user-specific bio-impedance signals from 3D motion and textual descriptions. It consists of two components: (1)Pose2Imp that simulates impedance dynamics from 3D body mesh sequences using shortest-path estimation, soft-body physics, and neural signal transformation.(2)Text2Imp that uses generative motion models to convert textual activity descriptions into pose sequences, then synthesizes corresponding impedance signals via Pose2Imp.
Unlike prior approaches that naively mix synthetic and real data, we propose \textbf{SImpHARNet}, a two-stage learning strategy designed to preserve domain alignment: (1) Contrastive Pretraining: Aligns simulated impedance and text embeddings to build semantically meaningful shared representations, and (2)Fine-tuning: Adapts the model to a target task using limited real-world impedance data.
We further introduce \textbf{ImpAct}, a new multimodal dataset of synchronized impedance, video, and 3D pose recordings across ten everyday activities and ten participants. This dataset fills a crucial gap for benchmarking impedance-based HAR models.

Our key contributions are:
\begin{itemize}
    \item \textbf{SImpHAR}, a simulation framework that generates realistic bio-impedance signals from motion and text, enabling scalable and controllable data synthesis.
    \item \textbf{SImpHARNet}, a decoupled training pipeline combining contrastive pretraining and fine-tuning for effective domain adaptation to real-world tasks.
    \item \textbf{ImpAct}, a new dataset for impedance-based HAR, offering aligned video, 3D pose, activity annotations, and wearable sensor data (\cref{fig:dataset}).
\end{itemize}

\section{Related Work}
A primary challenge in impedance-based HAR is the lack of large, labeled datasets. As with other sensor modalities, collecting synchronized and annotated data is expensive and labor-intensive. Prior efforts have addressed this gap through sensor simulation and cross-modal representation learning.

\textit{Sensor Simulation:} Several simulation frameworks generate synthetic sensor data using video, motion capture, or text. IMUTube~\citep{kwon2020imutube} and Cromosim~\citep{hao2022cromosim} synthesize inertial signals from pose data, while PresSim~\citep{ray2023pressim} and PressureTransferNet~\citep{ray2023pressuretransfernet} produce pressure maps from video. Capafoldable~\citep{ray2023capafoldable,ray2024origami} applies a similar approach to simulate capacitive signals for smart textiles.
Recent methods extend this idea by coupling text-to-motion generation with simulators, enabling sensor synthesis from natural language prompts. IMUGPT~\citep{leng2024imugpt} and T2P~\citep{10503379} generate inertial or pressure data using IMUSim~\citep{young2011imusim} and PresSim. While effective, these pipelines often struggle with domain-specific activity coverage due to the limited motion diversity of generative models. Moreover, mixing synthetic and real data without alignment can lead to performance degradation. Our method addresses this by decoupling representation learning from task fine-tuning: SImpHARNet uses contrastive pretraining on large-scale simulated data followed by adaptation on small, labeled real-world samples.

\textit{Representation Learning:} Contrastive and self-supervised learning have shown strong results in HAR by aligning representations across heterogeneous modalities~\citep{tang2020exploring,fortes2022learning,Deldari_2021}. IMU2CLIP~\citep{moon2022imu2clip} projects IMU signals into shared embeddings with images and text via CLIP~\citep{radford2021learning}. iMove~\citep{liu2024imove} jointly embeds IMU and impedance signals, while others~\citep{lago2,nguyen2023virtual,lago} demonstrate the benefits of training with richer sensor or auxiliary modalities.
Building on these insights, we simulate impedance signals at scale using motion and text inputs, then use contrastive learning to align them with corresponding descriptions. This enables SImpHARNet to generalize across activity classes—even when synthetic and real-world categories are not perfectly aligned—making it effective in data-scarce regimes.

\section{Approach}

The proposed approach, SImpHAR, addresses the data scarcity and modeling challenges in impedance-based HAR by introducing a simulation-to-training pipeline that can generate realistic, personalized impedance signals from motion data and support robust downstream learning. At the core is a simulation framework called Pose2Imp, which models impedance dynamics from 3D human pose using shortest-path analysis, soft-body deformation, and a neural grounding module that personalizes the signals based on physiological traits. 
Text2Imp complements this by leveraging generative text-to-motion models to produce diverse activity sequences from textual prompts, which are then converted into impedance signals via Pose2Imp. 
To overcome the limitations of Text-to-motion generative models in generating domain specific actions, we introduce SImpHARNet, a two-stage learning strategy that first performs contrastive pretraining on large-scale simulated impedance-text pairs to learn generalizable embeddings, and then fine-tunes the model on a small set of real, labeled signals. This decoupled design allows the model to benefit from diverse, unlabeled motion data during pretraining while ensuring adaptation to the target domain during fine-tuning.

\subsection{Pose2Imp}

Our method is based on the observation that bio-impedance is influenced by the path of least resistance (i.e., the shortest conductive path) between two electrodes on the body. As the body moves, this path dynamically changes—making it a proxy for estimating time-varying impedance signals.

\paragraph{Shortest Path Estimation.} 

\begin{figure}[!t]
\centering
\includegraphics[width=0.9\linewidth]{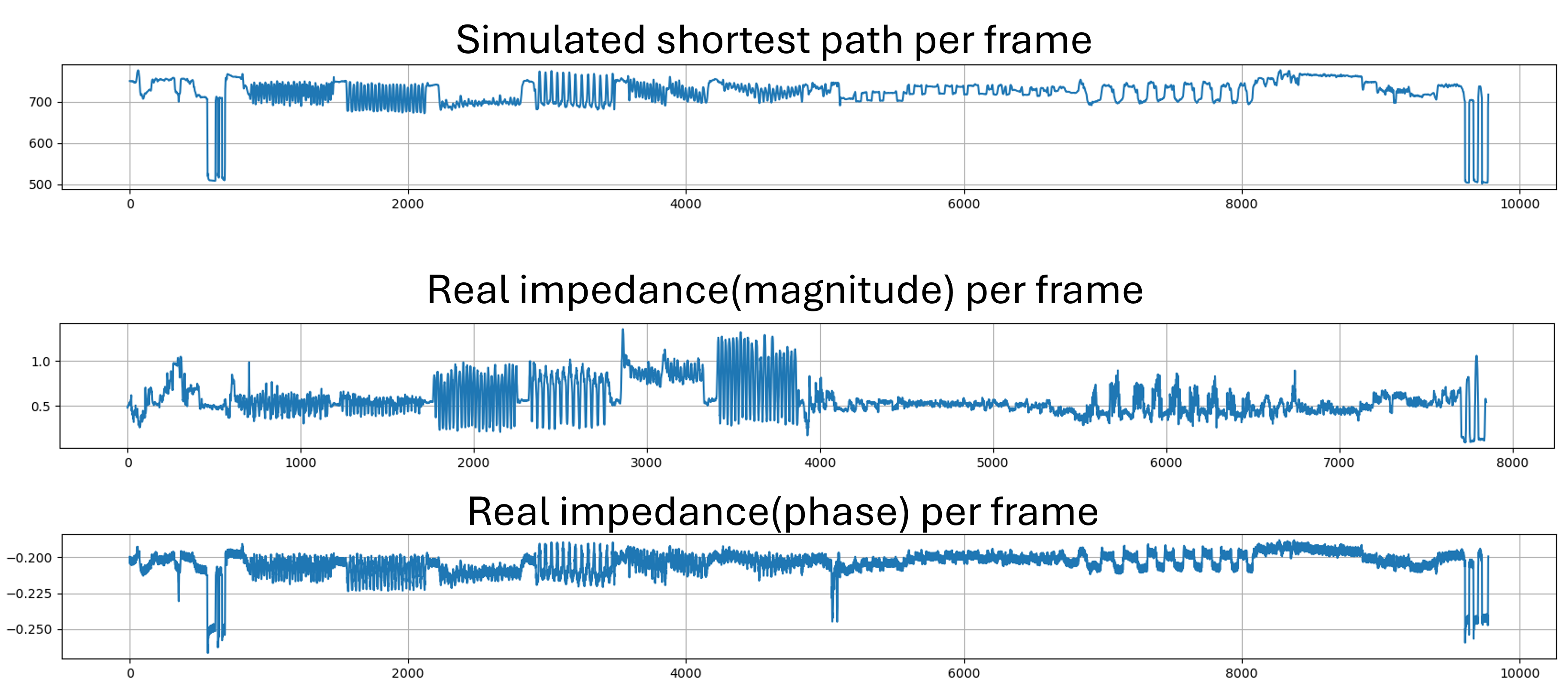}
\caption{Direct correlation between impedance magnitude, phase, and simulated shortest path.}
\label{fig:sim_v_real}
\end{figure}

We use OSX~\citep{lin2023one} to estimate SMPL-H~\citep{loper2023smpl} meshes from video. The body mesh is modeled as a graph where vertices represent surface points and edge weights correspond to Euclidean distances. 

We adapt Dijkstra's algorithm~\citep{zimmer2013efficient} to compute frame-wise shortest paths between electrode positions on the 3D body mesh. Unlike the standard algorithm that minimizes hop count, we assign edge weights using Euclidean distances between mesh vertices—yielding geodesic paths that approximate conductive paths relevant to bio-impedance.

\begin{algorithm}
\caption{Dijkstra’s Algorithm with Euclidean Weights on Mesh}
\label{euclid}
\begin{algorithmic}[1]
\footnotesize
\State $d[v] \gets \infty$ for all vertices $v$, except source $s$: $d[s] \gets 0$
\State Initialize priority queue $Q$ with $s$
\While{$Q$ not empty}
    \State Extract $u$ with minimum $d[u]$
    \For{each neighbor $v$ of $u$}
        \State $d[v] \gets \min(d[v], d[u] + \text{dist}(u, v))$
        \If{$d[v]$ updated}
            \State Update $v$ in $Q$
        \EndIf
    \EndFor
\EndWhile
\end{algorithmic}
\end{algorithm}

This process yields a time-series of shortest path lengths between electrodes, which serve as a geometric approximation of the underlying impedance signal.

\paragraph{Soft Body Simulation.}  
While shortest paths offer a useful structural estimate, they can be biased by mesh discretization and surface topology. To improve realism, we embed a deformable soft-body along the computed path, treating the SMPL mesh as a rigid scaffold. The soft body is initialized with the shortest-path vertices and then deformed using internal strain dynamics to produce a smoothed, continuous path. This step decouples the path shape from mesh resolution and results in more physiologically plausible impedance variation, as shown in \cref{fig:sim_v_real}.

\paragraph{User-Specific Grounding via Neural Mapping.}  
Simulated paths alone do not capture individual physiological factors such as body composition, hydration level, or muscle-to-fat ratio—all of which significantly influence real impedance measurements. To bridge this gap, we introduce a neural mapping module that calibrates the simulated signal to user-specific impedance dynamics.

\begin{figure}[!t]
\centering
\includegraphics[width=0.9\linewidth]{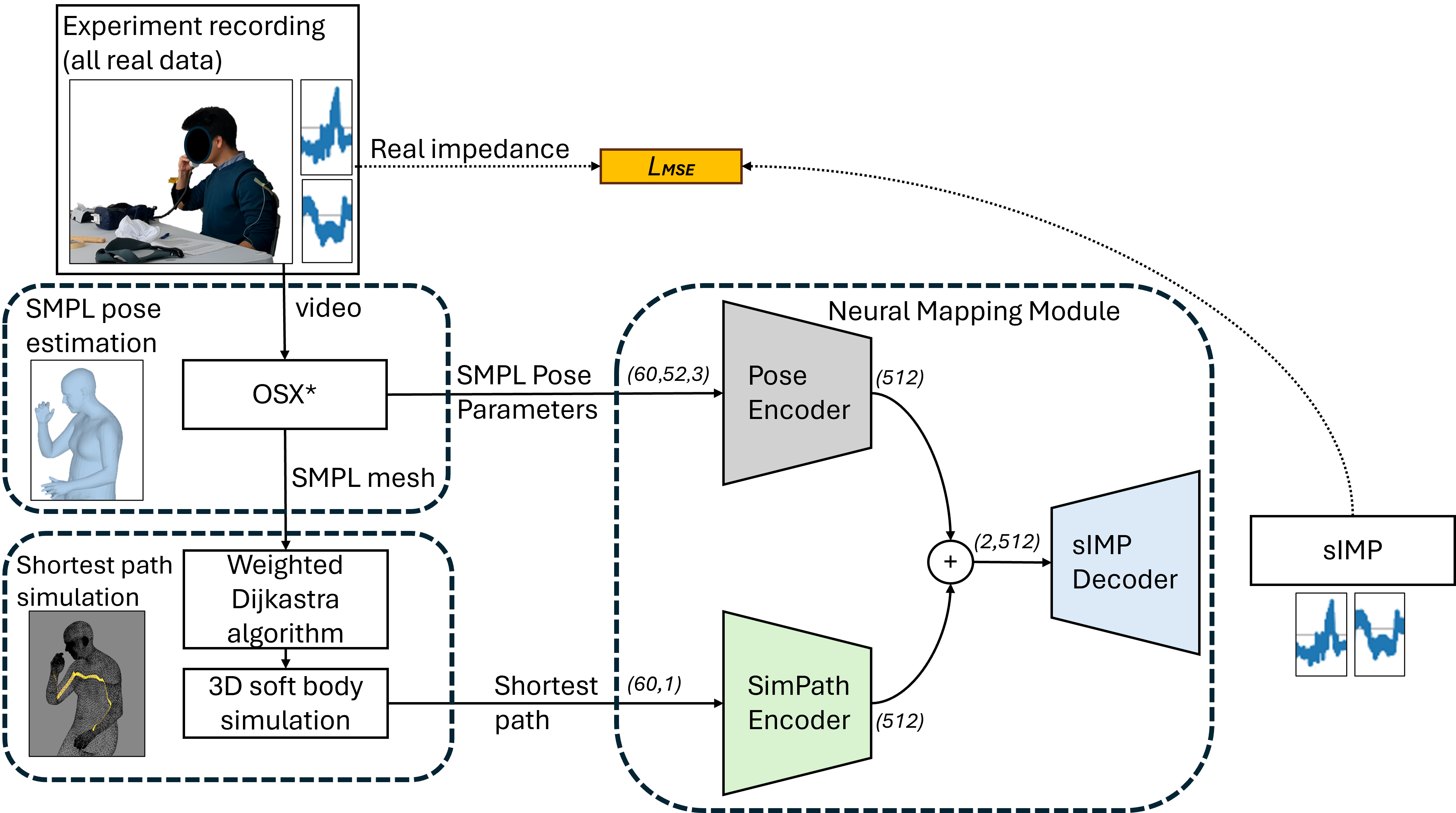}
\caption{Pose2Imp simulation involves shortest path estimation, followed by soft body simulation and user-specific grounding through neural mapping.}
\label{fig:pose2imp}
\end{figure}

We adopt a dual-encoder architecture inspired by PresSim~\citep{ray2023pressim}, illustrated in \cref{fig:pose2imp}. The network takes as input both the geometric simulation and the user’s pose, and outputs a calibrated impedance signal:

\begin{itemize}
  \item \textbf{SImpath Encoder:} A bidirectional LSTM that encodes the sequence of simulated path lengths with shape \texttt{(window, 1)}, where \texttt{window} refers to a temporal segment of 60 frames (2 seconds).
  \item \textbf{Pose Encoder:} A parallel bidirectional LSTM that processes SMPL pose parameters with shape \texttt{(window, 52, 3)} corresponding to 52 joints and 3 rotation axes.
  \item \textbf{CNN Decoder:} A convolutional decoder that fuses the two 512-dimensional encoder outputs and predicts the impedance magnitude and phase as a vector of shape \texttt{(1, 2)}.
\end{itemize}

The model is trained using paired real-world impedance and pose data with a Mean Squared Error (MSE) loss. This supervised grounding step enables the framework to personalize the simulated signals—ensuring that the output better reflects actual sensor behavior for a given individual.

\subsection{Text2Imp}  

\begin{figure}[!t]
\centering
\includegraphics[width=0.9\linewidth]{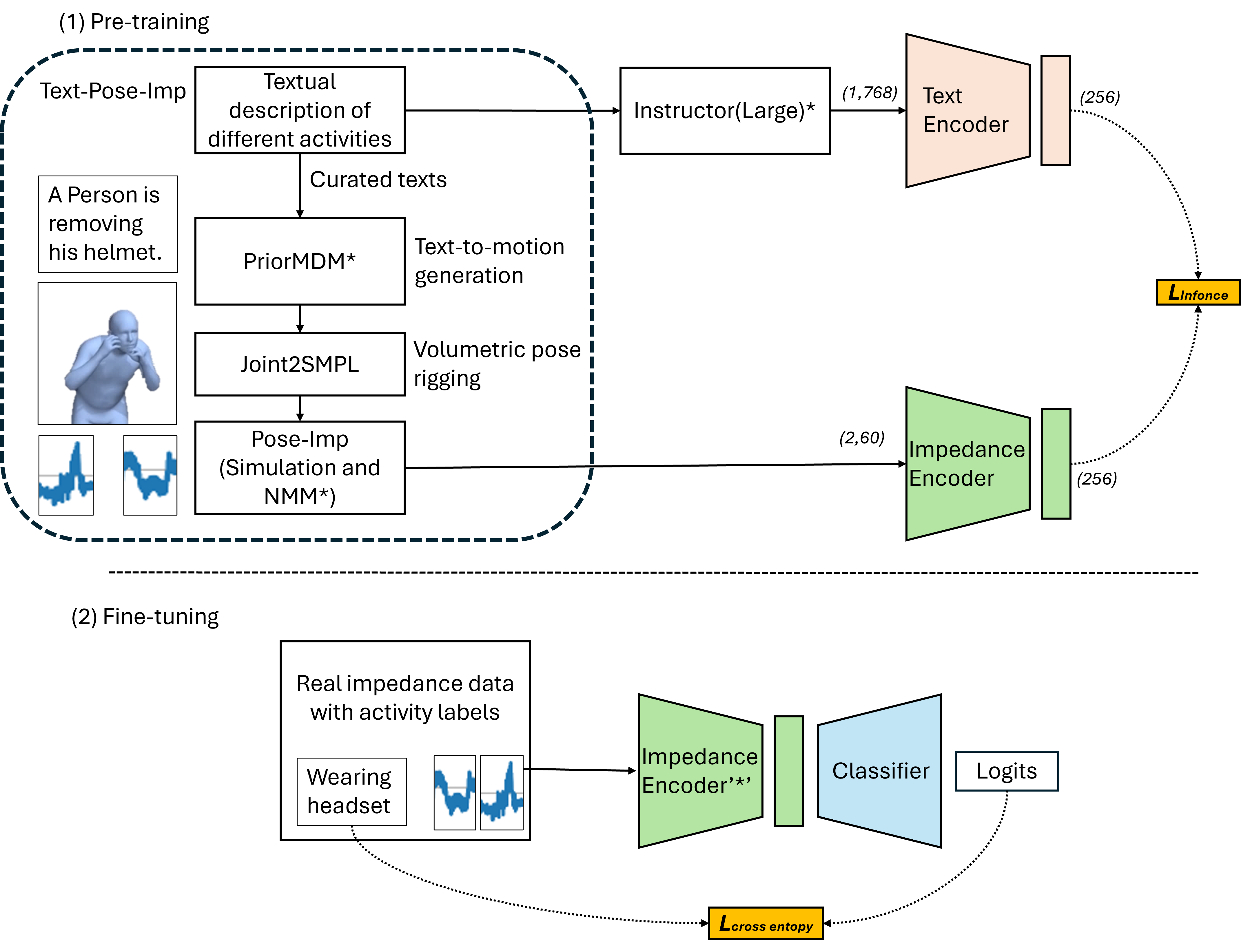}
\caption{SImpHARNet consists of 2 steps (1) pretraining with simulated data and text embedding for joint representation and (2) downstream classifier training using the pretrained Impedance encoder for HAR ("*" depicts pretrained models).}
\label{fig:overview}
\end{figure}

Manually collecting and labeling real-world videos that resemble target activities is slow, labor-intensive, and often requires physical supervision. As an alternative, text-to-motion generative models offer a scalable way to produce 3D motion data from textual descriptions. However, these models are inherently limited by the motion diversity of their training datasets. For example, common datasets like HumanML3D~\citep{guo2022generating} and KIT-ML~\citep{plappert2016kit} do not include actions such as "wearing VR glasses," making it impossible for the models to generate such motions directly.

To address this limitation, we leverage the generative flexibility of PriorMDM~\citep{shafir2023human}, a diffusion-based text-to-motion model, to synthesize motions that are semantically similar to our target activities. For each class in our dataset, we craft three related textual descriptions that reflect a similar motion range (e.g., "removing a helmet", "touching both ears" etc as a proxy for "wearing a VR headset") and generate at least ten variations per description, resulting in 30 motion samples per class.

The generated 3D pose sequences are converted into SMPL meshes using joint2SMPL and passed through our Pose2Imp simulation pipeline to produce synthetic impedance signals (SImp). While the generated motions do not exactly replicate those in our real dataset, they capture structurally similar dynamics. As shown in \cref{fig:tsne}, t-SNE visualization~\citep{van2008visualizing} of activity text embeddings illustrates strong alignment between real and synthetic samples, validating the semantic consistency of our approach.

\subsection{SImpHARNet}

\begin{figure}[!t]
\centering
\includegraphics[width=0.8\linewidth]{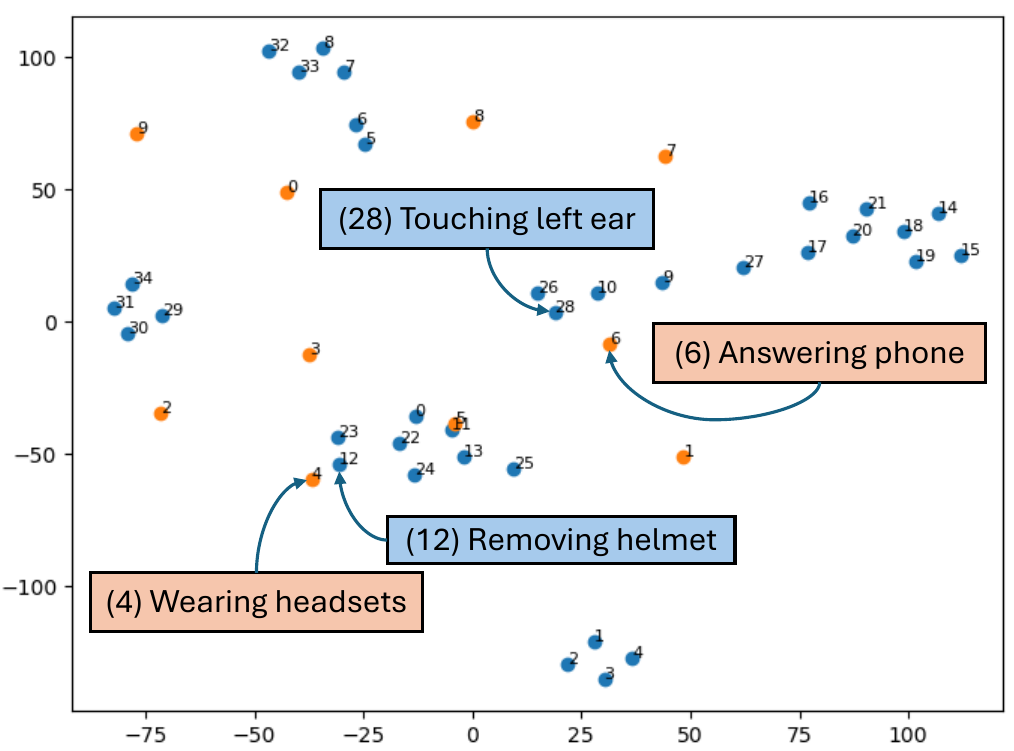}
\caption{Text embeddings of target classes (orange) and pretraining classes used for generating SImp signals (blue).}
\label{fig:tsne}
\end{figure}

To address the limitations of generative models—which often fail to produce domain-specific activities (e.g., wearing a VR headset)—and to avoid performance degradation from naïvely mixing real and synthetic data, we adopt a two-stage training strategy. As visualized in \cref{fig:overview}, we first pretrain using bi-modal contrastive learning on synthetic impedance-text pairs, followed by fine-tuning on a small set of labeled real-world data.

\paragraph{Pretraining.}
The first stage trains our contrastive representation learning model, \textbf{SImpHARNet}, to align synthetic impedance signals (SImp) with their associated textual prompts. The goal is to create a shared embedding space where semantically related pairs are close together, enabling generalization to unseen or low-resource activity classes.

SImpHARNet consists of two parallel encoders:
\begin{itemize}
    \item A fully connected (FC) text encoder with three linear layers, batch normalization, dropout, and GeLU activations.
    \item A DeepConvLSTM-inspired impedance encoder that transforms a 60-frame SImp window into a 256-dimensional feature vector.
\end{itemize}

Text prompts are first encoded using Instructor (Large)~\citep{su2022one}, a pre-trained language model that produces a 1,768-dimensional sentence embedding. Each contrastive training pair consists of a text prompt and its corresponding synthetic impedance signal.

Matched pairs are generated by feeding a specific prompt (e.g., \textit{“removing a helmet”}) into our Text2Imp pipeline and pairing the resulting SImp signal with the original prompt.  
Unmatched pairs are sampled by randomly selecting text prompts that describe semantically different activities (e.g., pairing “removing a helmet” with SImp from “answering a phone”).

We train the encoders using the InfoNCE loss, which minimizes the distance between matched (positive) pairs while maximizing separation from unmatched (negative) ones:
\begin{equation}
 \mathcal{L}_\text{InfoNCE}(q, k) = -\frac{1}{N} \sum_{i=1}^{N} \log \left( \frac{e^{s(q_i, k_i)}}{e^{s(q_i, k_i)} + \sum_{j \neq i} e^{s(q_i, k_j)}} \right)
\end{equation}
where similarity is computed using cosine distance:
\begin{equation}
 s(q, k) = \frac{q \cdot k}{\lVert q \rVert \cdot \lVert k \rVert}
\end{equation}

To support semantic alignment, we manually craft three text prompts per target activity class (e.g., \textit{“removing helmet”}, \textit{“touching both ears”}, \textit{“putting on a hat”} for the action \textit{“wearing VR headset”}), and use PriorMDM to generate multiple pose sequences per prompt. These are then converted into impedance signals via the Pose2Imp simulation pipeline.

We validate the resulting semantic clusters via t-SNE~\citep{van2008visualizing}, showing that related real and synthetic activity classes form tight clusters in the embedding space (\cref{fig:tsne}).

\paragraph{Downstream Training.}
In the second stage, we fine-tune the impedance encoder for the target HAR task using real-world labeled data. A lightweight classification head—consisting of a 1D CNN followed by two fully connected layers—is trained on top of the frozen impedance encoder.

To avoid catastrophic forgetting, the impedance encoder is initially frozen during the first $N$ epochs. Once the classification loss plateaus, we unfreeze the encoder and continue fine-tuning the full model end-to-end. This staged optimization ensures that pretrained semantic representations are preserved while allowing adaptation to the target domain.

\section{Experimental Results}
\subsection{Datasets}
\label{data}

\paragraph{ImpAct Dataset}
Ten participants (four women and six men), aged 24 to 35 years, with body weights ranging from 50.6 to 88.15 kg and heights between 160 and 188 cm, participated in the study. Each participant performed a total of eight activities: six upper-body fitness exercises and three daily tasks involving fine hand movements. Each session lasted approximately 15 minutes per participant.
Participants wore a custom-designed bioimpedance sensing device comprising four components: an analog front-end (AFE), a control module, surface electrodes, and a power supply following the architecture in \citet{liu2024imove}. The AFE used the AD5941 chip (Analog Devices) to generate sine-wave voltage stimuli across 0.015 Hz to 200 kHz and capture the resulting current with a high-speed transimpedance amplifier. The control module, based on Nordic's nRF52840 SoC, communicated with the AFE via SPI and wirelessly streamed the measurements over Bluetooth.
Bioimpedance electrodes were placed on both wrists one acting as the transmitter and the other as the receiver forming a closed electrical loop with a shared ground. This configuration allowed impedance to be measured across the upper body in a wearable, non-intrusive setup. Additionally, an IMU was mounted on the dominant wrist for motion tracking.
The six fitness exercises included \textit{Box}, \textit{Biceps curl}, \textit{Chest press}, \textit{Shoulder and chest press}, \textit{Arm hold and shoulder press}, and \textit{Arm opener}. The three daily tasks were \textit{Sweeping a table}, \textit{Ringing a headphone}, and \textit{Transferring a telephone}, focusing on hand and arm manipulation.
Bioimpedance and IMU data were streamed in real time to a JavaScript-based laptop application that supported live visualization and automatic text-based logging. A rear-facing video camera recorded all sessions for manual annotation. After temporal synchronization, all data streams were sampled at 20 Hz and segmented using a sliding window of 50 samples with a step size of 10. A photo of the experimental setup, showing electrode placement and device configuration.

\paragraph{iMove Dataset}
The iMove dataset contains synchronized multimodal sensor data collected from ten participants (four female, six male), aged 24 to 35 years. Each participant performed six upper-body fitness exercises—\textit{Box}, \textit{Biceps Curl}, \textit{Chest Press}, \textit{Shoulder and Chest Press}, \textit{Arm Hold}, and \textit{Arm Opener}—across five consecutive days. The dataset includes bio-impedance signals from both wrists and IMU data from the left wrist. All signals were sampled at 20 Hz and segmented using a sliding window of 50 samples with a step size of 10.

\paragraph{iEat Dataset}
The iEat dataset comprises synchronized multimodal data collected from ten participants, each completing 40 meal sessions in naturalistic dining settings. Participants engaged in a diverse set of dietary-related activities, including \textit{eating with utensils}, \textit{drinking}, \textit{using napkins}, \textit{talking}, and \textit{reaching for condiments}, among others. The dataset captures bio-impedance signals from both wrists using a wearable device, sampled at 20 Hz. 

\subsection{Training Details}
The model is written in PyTorch and trained on a linux device with CUDA support, while the simulation is written in Python using the Blender API.
We used 300 epochs and employed early stopping with 15 epochs, ADAMW optimization, and a learning rate of 0.001, the ReduceLR On Plateau, with patience 10.
We used a leave-one-user-out cross-validation paradigm to evaluate our model on all three datasets.
During training, the training and validation data are split with a 9:1 ratio based on a random split.

\subsection{Results}

\begin{table}[!t]
\footnotesize
  \caption{Activity classification results for baseline, TinyHAR \citep{zhou2022tinyhar}, M2L \citep{yang2022more}, SR \citep{liu2024imove}, TCNNet \citep{musallam2021electroencephalography} and SImpHARNet (ours) using different strategies. For the iMove and iEat datasets, the given accuracy using M2L, SR, TCNNet, and TinyHAR is reported in the respective dataset papers \citep{liu2023ieat, liu2024imove}.}
  \label{tab:res}
  \centering
  \resizebox{\linewidth}{!}{
  \begin{tabular}{lccc}
    \toprule
    Dataset & Model & Accuracy & Macro F1\\
    \midrule
    \multirow{8}{*}{ImpAct} 
    & Baseline classifier (real data) & 0.621$\pm$0.011 & 0.595$\pm$0.007 \\
    \cmidrule(lr){2-4}
    & Baseline classifier (real+SImp data) & 0.618$\pm$0.016 & 0.581$\pm$0.012 \\
    & TinyHAR (real+SImp data) & 0.633$\pm$0.015 & 0.608$\pm$0.011 \\
    & M2L (real+SImp data) & 0.655$\pm$0.014 & 0.618$\pm$0.012 \\
    & SR (real+SImp data) & 0.661$\pm$0.013 & 0.633$\pm$0.011 \\
    & TCNNet (real+SImp data) & 0.675$\pm$0.012 & 0.648$\pm$0.011 \\
    \cmidrule{2-4}
    & SImpHARNet (frozen) & 0.651$\pm$0.015 & 0.612$\pm$0.009 \\
    & SImpHARNet (fine-tuned) & \textbf{0.775$\pm$0.010} & \textbf{0.767$\pm$0.012} \\
    \midrule
    \multirow{8}{*}{iMove}
    & Baseline classifier (real data) & 0.641$\pm$0.012 & 0.636$\pm$0.008 \\
    \cmidrule(lr){2-4}
    & Baseline classifier (real+SImp data) & 0.743$\pm$0.014 & 0.739$\pm$0.010 \\
    & TinyHAR (real+SImp data) & 0.755$\pm$0.013 & 0.748$\pm$0.012 \\
    & M2L (real+SImp data) & 0.762$\pm$0.012 & 0.754$\pm$0.011 \\
    & SR (real+SImp data) & 0.770$\pm$0.013 & 0.764$\pm$0.012 \\
    & TCNNet (real+SImp data) & \textbf{0.778$\pm$0.012} & \textbf{0.771$\pm$0.010} \\
    \cmidrule(lr){2-4}
    & SImpHARNet (frozen) & 0.704$\pm$0.017 & 0.696$\pm$0.011 \\
    & SImpHARNet (fine-tuned) & 0.763$\pm$0.016 & 0.752$\pm$0.015 \\
    \midrule
    \multirow{8}{*}{iEat}
    & Baseline classifier (real data) & 0.731$\pm$0.012 & 0.718$\pm$0.020 \\
    \cmidrule(lr){2-4}
    & Baseline classifier (real+SImp data) & 0.669$\pm$0.015 & 0.651$\pm$0.013 \\ 
    & TinyHAR (real+SImp data) & 0.650$\pm$0.014 & 0.643$\pm$0.012 \\
    & M2L (real+SImp data) & 0.665$\pm$0.013 & 0.658$\pm$0.011 \\
    & SR (real+SImp data) & 0.680$\pm$0.014 & 0.673$\pm$0.012 \\
    & TCNNet (real+SImp data) & 0.690$\pm$0.013 & 0.683$\pm$0.011 \\
    \cmidrule(lr){2-4}
    & SImpHARNet (frozen) & 0.791$\pm$0.008 & 0.786$\pm$0.012 \\
    & SImpHARNet (fine-tuned) & \textbf{0.844$\pm$0.014} & \textbf{0.832$\pm$0.009} \\
  \bottomrule
\end{tabular}
}
\end{table}

\begin{table}[!t]
\footnotesize
  \caption{Performance of STN trained on different inputs.}
  \label{tab:synth}
  \centering
  \begin{tabular}{lc}
    \toprule
    TSN Input &$R^2$\\
    \midrule
    3D Pose & 0.414$\pm$0.018 \\
    Simulated Distance & 0.821$\pm$0.013 \\
    3D Pose + Simulated Distance & \textbf{0.843$\pm$0.025} \\
  \bottomrule
\end{tabular}
\end{table}

\begin{figure}[!t]
\centering
\includegraphics[width=0.8\linewidth]{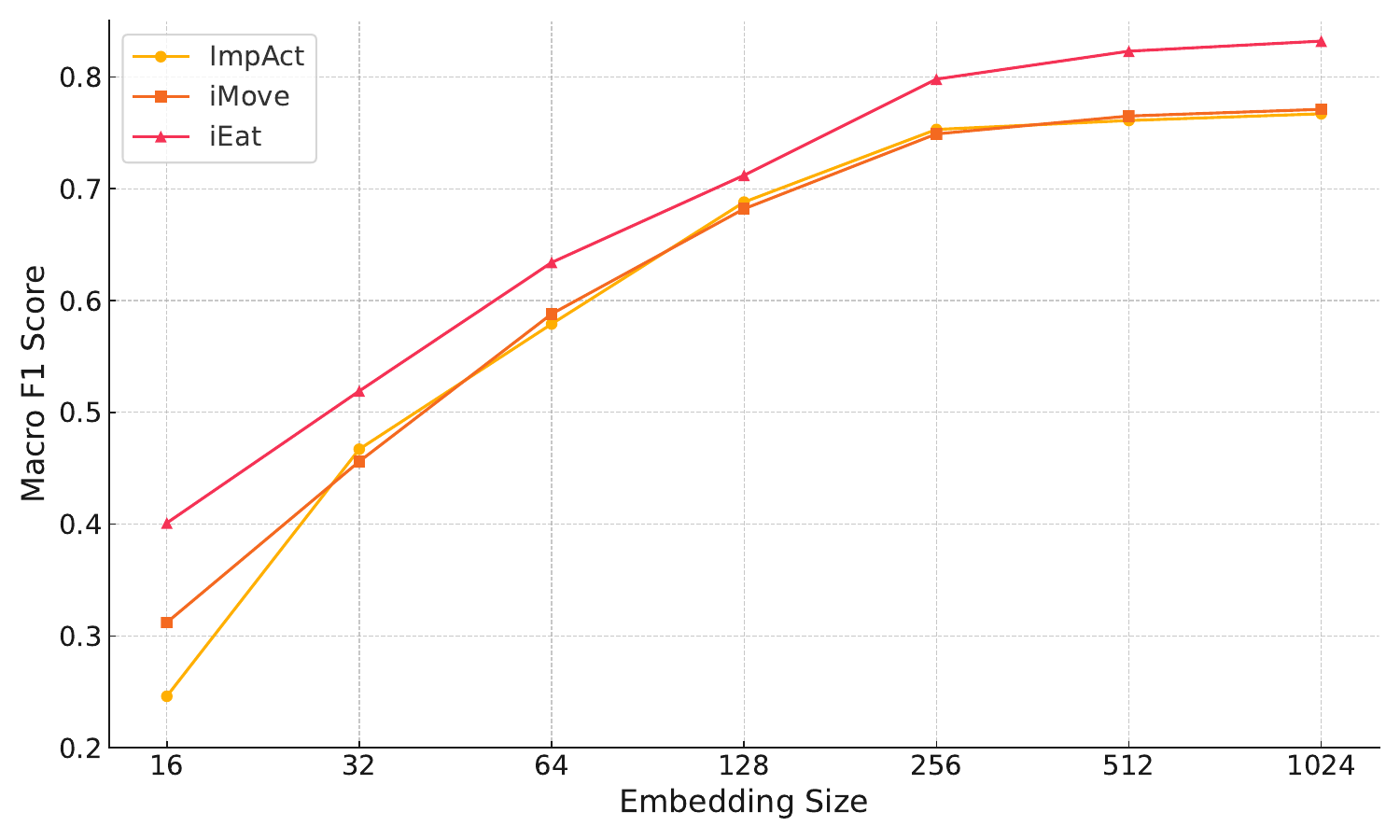}
\caption{SImpHARNet trained with varying embedding size showcasing the change in F1 based on dimension size.}
\label{fig:dim}
\end{figure}

\begin{figure}[!t]
\centering
\includegraphics[width=0.8\linewidth]{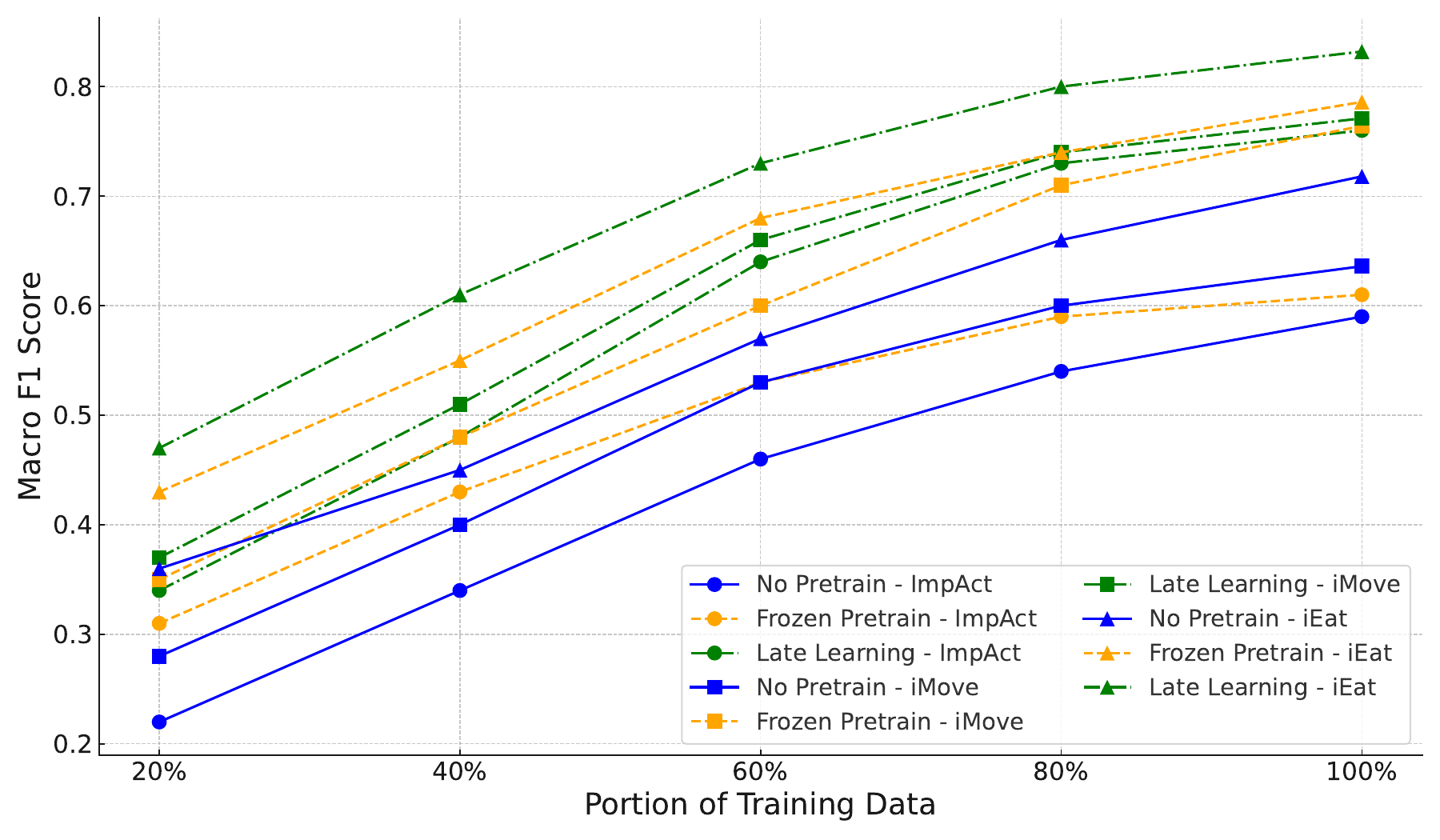}
\caption{Change in F1 score for classification models trained with different portions of the training set.}
\label{fig:data}
\end{figure}

We trained all models five times and report the mean R$^2$ for SImpHAR's neural mapping given in \cref{tab:synth} and macro F1/accuracy for downstream HAR tasks using SImpHARNet given in \cref{tab:res}. 

\textit{Effectiveness of Simulated Shortest Path: }To evaluate the benefit of our geometric simulation, we trained the STN using only SMPL pose parameters, which yielded an R$^2$ of 0.414. Incorporating simulated shortest path distances improved performance by nearly 50\%, and combining both inputs yielded the best result (R$^2$ = 0.843), validating the contribution of the path-based simulation. 

\textit{Impact of Simulated Impedance Data on HAR: }We evaluated the effect of augmenting real data with synthetic impedance (SImp) signals for HAR. Across datasets, adding SImp data to real training samples yielded modest improvements in some cases. On the \textit{iMove} dataset, where PriorMDM can accurately generate most activity classes, macro F1 increased from 0.636 to 0.739, representing a relative gain of 16.2\%. However, performance on \textit{iEat} declined from 0.718 to 0.651, likely due to a semantic mismatch between generated and real-world activities. On \textit{ImpAct}, there is a loss of performance as well. These results highlight the potential but also the limitations of naïvely mixing synthetic and real data, underscoring the need for more structured integration strategies such as those used in SImpHARNet. 

\textit{SImpHARNet Architecture Performance: }SImpHARNet’s two-stage design—contrastive pretraining followed by task-specific fine tuning enables robust representation learning from synthetic data and effective adaptation to real-world HAR. This architecture consistently outperforms both baseline and single-stage models across \textit{ImpAct}, \textit{iMove}, and \textit{iEat}, particularly in low-data regimes. The encoder learns transferable features during pretraining, while the downstream classifier benefits from flexibility during fine-tuning, resulting in substantial gains in both macro F1 and accuracy. 

\textit{Encoder Freezing vs. Late Learning: }We compared two training regimes: (1) freezing the encoder, and (2) staged fine-tuning (Late Learning). The latter consistently yielded higher F1 scores (0.767 for ImpAct, 0.771 for iMove, 0.832 for iEat). Frozen models showed flatter performance curves across training data sizes, reflecting limited adaptability due to fixed encoder weights. In contrast, Late Learning leveraged additional data more effectively. 

\textit{Embedding Dimension: }We evaluated how bottleneck dimensionality affects performance given in \cref{fig:dim}. Increasing embedding size from 16 to 256 improved F1 scores across all datasets, as larger vectors captured richer motion features. Gains plateaued beyond 256, with marginal improvement up to 1024. Peak F1 scores ranged from 0.753 to 0.832 at 256 or 512 dimensions. We chose 256 as a balance between accuracy and efficiency. 

\textit{Training Data Size: }To test data efficiency, we varied the training set size from 20\% to 100\% given in \cref{fig:data}. Across all datasets, Late Learning surpassed other strategies even with 60–80\% of data. For instance, in iEat, it outperformed the full-data No Pretrain baseline using only 60\%. No Pretrain improved slowly with more data, while Frozen Pretrain was stable but less responsive, further validating the value of our staged pretraining approach.

\section{Limitations}
While our framework demonstrates strong performance across multiple datasets, it is subject to two significant limitations that highlight directions for future improvement. (1) \textit{Limited Biophysical Fidelity}: The current framework does not explicitly model individual physiological factors such as tissue conductivity, hydration levels, or body composition, which can significantly influence real-world impedance signals. Additionally, the neural grounding model requires subject-specific impedance-pose data for calibration, limiting the framework’s ability to generalize across users without retraining.
Integrating biophysical priors into the simulation process could enhance physiological realism. To reduce calibration overhead, few-shot learning, meta-learning, or conditioning on user metadata (e.g., height, weight, body type) may enable signal personalization without subject-specific ground-truth data.
(2) \textit{Evaluation Scope}: Our experiments are focused on upper-body activities using wrist-mounted electrodes and a relatively small, demographically narrow participant pool. As a result, generalizability to lower-body movements, alternative electrode placements, or more diverse populations remains untested.
Expanding evaluations to include additional activity types, sensor locations (e.g., chest, thigh), and varied participant demographics can provide a broader assessment. Cross-dataset validation would further support claims of robustness.

\section{Conclusion}
We presented SImpHAR, the first simulation framework for generating realistic bio-impedance signals from 3D motion and text prompts, enabling scalable data synthesis for human activity recognition. Central to our pipeline is Pose2Imp, a physics-informed method that models impedance dynamics via shortest paths and soft-body deformation, and Text2Imp, which bridges language and motion using generative models. We introduced SImpHARNet, a two-stage training strategy that leverages contrastive pretraining on synthetic impedance-text pairs, followed by fine-tuning on limited real data.
Extensive experiments on three Impedance based HAR datasets ImpAct, iMove, and iEat—demonstrated that SImpHARNet consistently outperforms state-of-the-art baselines, particularly in low-data regimes. Our ablations confirm the effectiveness of simulated impedance, contrastive objectives, and staged fine-tuning.
Looking ahead, we aim to integrate physiological models into the simulation pipeline, support broader activity coverage and sensor placements, and explore domain adaptation techniques to enable zero-calibration deployment across users.


\begin{acks}
The research reported in this paper was supported by the BMBF in the project CrossAct (01IW21003) and IITP grant funded by the Korea government(MSIT) (No. RS-2019-II190079).
\end{acks}
\bibliographystyle{ACM-Reference-Format}
\bibliography{sample-base}


\end{document}